\documentclass{article}
\PassOptionsToPackage{prologue,dvipsnames}{xcolor}

\usepackage{arxiv}
\usepackage{mathpazo}
\usepackage{hyperref}
\usepackage{url}

\usepackage[round]{natbib}
\usepackage{xspace}
\usepackage{amsmath, bm}
\usepackage{listings}
\usepackage{mathtools}
\usepackage{siunitx}
\usepackage{xspace}
\usepackage{ntheorem}
\usepackage{graphicx}
\usepackage{booktabs}
\usepackage{bbm}
\usepackage{stmaryrd}
\usepackage{algorithm}
\usepackage{algorithmic}
\usepackage{nth}
\usepackage{subcaption}
\usepackage{tikz}
\usepackage{pgfplots}
\usepackage{setspace}

\author{%
  Yongchang Hao$^\spadesuit$\thanks{Project done during Mitacs internship at Borealis AI.} \quad Yanshuai Cao$^\diamondsuit$ \quad Lili Mou$^{\spadesuit\,\heartsuit}$\\
  $^\spadesuit$Dept. Computing Science \& Alberta Machine Intelligence Institute (Amii), University of Alberta \\
  $^\diamondsuit$Borealis AI \quad 
  $^\heartsuit$Canada CIFAR AI Chair \\
 \texttt{yongcha1@ualberta.ca}\\\texttt{yanshuai.cao@borealisai.com}\quad\texttt{doublepower.mou@gmail.com}
}

\newcommand{\name}{NeuZip\xspace}

\newcommand{\mytitle}{\name: Memory-Efficient Training and Inference with Dynamic Compression of Neural Networks}
\title{\mytitle}

\usepackage{amsmath,amsfonts,bm,xspace,multirow}
\definecolor{myred}{RGB}{255,127,127}
\definecolor{mygreen}{RGB}{127,255,127}

\def\1{\bm{1}}

\DeclareMathAlphabet{\mathsfit}{\encodingdefault}{\sfdefault}{m}{sl}
\SetMathAlphabet{\mathsfit}{bold}{\encodingdefault}{\sfdefault}{bx}{n}

\theoremstyle{plain}
\newtheorem{theorem}{Theorem}[section]

\newtheorem{observation}[theorem]{Observation}
\begin{document}

\maketitle

\begin{abstract}
The performance of neural networks improves when more parameters are used. However, the model sizes are constrained by the available on-device memory during training and inference. Although applying techniques like quantization can alleviate the constraint, they suffer from performance degradation. In this work, we introduce \name, a new weight compression scheme based on the entropy of floating-point numbers in neural networks. With \name, we are able to achieve memory-efficient training and inference without sacrificing performance. Notably, we significantly reduce the memory footprint of training a Llama-3 8B model from 31GB to less than 16GB, while keeping the training dynamics fully unchanged. In inference, our method can reduce memory usage by more than half while maintaining near-lossless performance. Our code is publicly available.\footnote[1]{\url{https://github.com/BorealisAI/neuzip}}
\end{abstract}

\section{Introduction}\label{sec:intro}

Deep learning with neural networks has become the backbone of numerous artificial intelligence applications. The search for better-performing networks is a longstanding topic in deep learning. Without modifying the design, scaling up the number of parameters (e.g., number of hidden dimensions or layers) has been demonstrated as an effective practice to boost the performance of neural networks of the same kind~\citep{kaplan2020scaling}. This idea has been successfully applied to text, image, audio, and multi-modal tasks with a wide range of model architectures~\citep{yu2022scaling,radford2019language,brown2020language}. Recently, the number of parameters in the state-of-the-art models has become more than 100 billion or even a trillion parameters. For example, one of the state-of-the-art language models in 2020, GPT-3, has 175B parameters~\citep{brown2020language}, growing by nearly 100 times compared with the largest Transformer architecture in the 2017 paper~\citep{vaswani2017attention}.

Despite the growth in the model size, the hardware capacity is not keeping up with the pace: the largest on-device memory of GPUs was 32GB in 2017, and is 80GB to this date in 2024, growing by only 2.5 times. The available hardware supply poses a limitation on the trainable model size, bottlenecking the scaling capacity. Although this problem can be alleviated by using more GPUs and sharding the model in multiple devices~\citep{rajbhandari2019zero}, such a practice introduces more communication overheads among GPUs, making large-scale distributed training less efficient. Therefore, saving the total memory usage is critical in scaling up neural networks.

The peak memory usage is dominated by three relatively independent parts: the optimizer, the saved activations for back-propagation, and the model itself. For the optimizer, there are already memory-efficient optimizers achieving a sublinear space complexity~\citep{shazeer2018adafactor,hao2024flora}; for the activations, the memory can be saved by enabling activation checkpointing~\citep{chen2016training}, which saves the storage by recomputing the forward activations during the back-propagation. For the model parameters, there has not been an effective method to save the memory while preserving the ability to train the model. Recently, \cite{dettmers2023qlora} proposed the quantized low-rank adaptation (QLoRA), which freezes the parameters using a 4-bit data type for the backbone pre-trained model. While significantly saving the memory for the model, it imposes a constraint on the overall change of the model to be low-rank, limiting the model capacity.

In this paper, we propose \name, an algorithm to compress the neural networks while maintaining their full abilities. Specifically, each floating-point number is represented by three parts: the sign bit, the exponent bits, and the mantissa bits. Following the observation that weights are concentrated around zero~\citep{kalamkar2019study}, we demonstrate that this corresponds to the low-entropy nature of the exponent bits. We hence compress the exponent bits using the asymmetric numeral system (ANS~\cite{duda2013asymmetric}), a lossless compression algorithm that achieves a high throughput on parallel computing devices like GPUs. Since the compression is lossless, the memory reduction comes without compromising any precision loss and enables full-parameter training.

In addition to lossless compression for training, we also propose a lossy variant of \name for inference that further reduces the memory footprint. Specifically, we control the relative change of each parameter by storing only the top-$k$ significant bits of the mantissa. We empirically show that lossy \name lies at the Pareto frontier of the memory--performance trade-off when compared with several state-of-the-art quantization baselines.

\section{Our Approach}\label{sec:approach}

The Shannon entropy~\citep{shannon1948mathematical} is used to measure the ``stochasticity'' of a random variable with the following definition:
\begin{align}
    H(X) := \mathop{\mathbb{E}}_{X\sim p(X)}[ -\log_2 p(X) ]
\end{align}
for a random variable $X$ with probability $p$.
A lower entropy indicates a less stochasticity of a random variable. In fact, the entropy equals the minimum number of bits required, in expectation, to represent a random variable, therefore corresponding to data compressibility. For the non-concentrating random variable with all possible values sharing an equal probability, the entropy of which reaches the maximum value $\log_2 n$, where $n$ is all possible values $X$ can take. On the other hand, for highly-concentrating (e.g., fully deterministic) random variables, the entropy can be as low as 0.

\subsection{Low-Entropy Nature of Neural Network Parameters}
We argue that the parameters in neural network tend to have low entropy. First, parameters are typically initialized with Gaussian distribution for matrices~\citep{glorot2010xavier,he2015delving}. This encourages all weights to be centered around zero, effectively reducing the entropy (or randomness). In addition, regularization is also applied for better generalization ability. For example, the weight decay technique reduces the magnitudes of weights at every update iteration. Similarly in Bayesian inference, prior distributions (e.g., Gaussian and Laplace distributions) are often applied, imposing a zero-concentrated preference over the parameters.Even without explicit regularization,
stochastic gradient descent (SGD) or its variants are shown to have the implicit regularization effect on neural networks, meaning the model parameters are implicitly encouraged to have smaller magnitudes during training~\citep{soudry2017implicit,vardi2021implicit}. 
All the above effects and techniques lead to the following observation:

\begin{observation}\label{obs:low-entropy}
Assuming neural network parameters are i.i.d.~random variables, the entropy of the distribution is likely to be low.
\end{observation}

Specifically, each parameter is represented is represented by three components: the sign bit, the exponent bits, and the mantissa bits in the IEEE 754 standard~\citep{ieee754}.\footnote{We use BF16~\citep{kalamkar2019study} in this paper.} Therefore, we conduct a fine-grained analysis and investigate the distribution of each component of a floating-point number in neural networks.

\begin{figure}[t]
    \centering
\includegraphics[width=\textwidth]{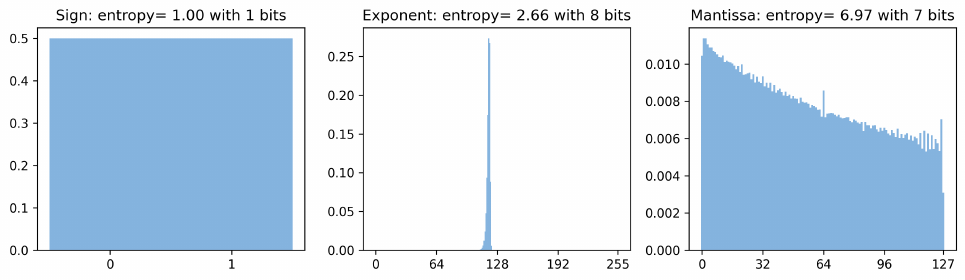}
    \caption{The histograms of different components of the parameters of LLama-3 8B model~\citep{dubey2024llama}. 
    The $x$-axis is all possible binary values and the $y$-axis represent the frequency of each value.}
    \label{fig:component-entropies}
\end{figure}

As shown in Figure~\ref{fig:component-entropies}, the sign bit has a high entropy as it is evenly distributed; hence, it is not compressible. For the exponent bits, there is a clear pattern that they demonstrate a low-entropy nature, carrying only less than 3 bits of information with 8 bits of capacity. For the mantissa bits, they store nearly 7-bit information with 7-bit capacity. In fact, we shown in Appendix~\ref{sec:apx:more_ent} that this is common in deep learning.

This phenomenon suggests that by simply compressing the exponents, we are able to recover the overall optimal compression ratio. In this example, an ideal compression algorithm is able to achieve a ratio as high as 1.501 (the sum of the three entropy values), only marginally below the overall compression ratio 1.505.

\subsection{Lossless \name: Compressing Exponents for Training}\label{sec:approach:lossless}
{

\paragraph{Compressed representation.} Based on observation, we see that the number of bits per exponent is largely inflated compared with the information entropy. However, previous research demonstrates that the dynamic range provided by the 8-bit exponents are critical for neural networks~\citep{kalamkar2019study}. We therefore propose to compress the exponent bits in a lossless manner based on the entropy. This practice mainly has three benefits:  (1) it increases the throughput of compression as only part of the bits are processed by the compression; (2) it reduces the burden of maintaining the statistics of a large set of symbols (e.g., 256 symbols for 8-bit exponents versus 65,536 symbols for 16-bit representations), enabling a great efficiency of compression algorithms; (3) most importantly, it recovers most of the compressibility as shown in Figure~\ref{fig:component-entropies}.

\paragraph{Multi-layer neural networks.} 
The compression alone does not save any memory for maintaining a single array. This is because, either compression or decompression, requires at least one buffer of the same size as the uncompressed array. In the scope of neural networks, the whole model is prohibitively large and it is infeasible to duplicate the memory. In \name, however, we exploit the multi-layer structure of modern neural networks to avoid creating a large buffer. Without loss of generality, we focus on the linear function as a common building block in neural networks at layer~$l$:
\begin{align}
    \bm{x}_{l} \gets  \bm{W}_l \bm{x}_{l-1} + \bm{b}_l,
\end{align}
where $\bm{W}_l \in \mathbb{R}^{m \times n}$ is the weight matrix, $\bm{b}_l \in \mathbb{R}^{m}$ is the bias vector of layer $l$, and $\bm{x}_l$ is the input of layer $l$. We propose to modify the compressed forward pass in the following form \begin{align}
    \hat{\bm{W}} \gets& \mathrm{decompress}(\bm{c}_l) \\
    \bm{x}_{l} \gets&  \hat{\bm{W}}\bm{x}_{l-1} + \bm{b}_l,
\end{align}
where $\bm{c}_l$ is the compressed storage of the matrix $\bm{W}_l$. In this way, we only need to store $\bm{c}_i$ for each layer, enjoying low-memory usage.  During each forward pass, weight matrices stay in the compressed form until the original data is needed, in which case it is decompressed into a temporary space $\hat{\bm{W}}$ for computation. As a result, the entire network is never fully decompressed at any point in time, making the overall forward pass memory efficient. The per-layer procedure is shown in Figure~\ref{fig:autograd}.

Note that although we alter the forward pass, the back-propagation for each linear layer is fully unaffected. This is because \begin{align}
    \frac{\partial \mathcal{L}}{\partial \bm{W}_l} = \frac{\partial \mathcal{L}}{\partial\bm{x}_l } \frac{\partial\bm{x}_l}{\partial \bm{W}_l} = \left(\nabla_{\bm{x}_l} \mathcal{L} \right)\bm{x}_{l-1}^\top. \label{eq:linear-gradient}
\end{align}
Therefore, we are able to obtain the gradient as long as the activations are saved. Similarly, we can also propagate the gradient of inputs with \begin{align}
        \frac{\partial \mathcal{L}}{\partial \bm{x}_{l-1}} = \frac{\partial \mathcal{L}}{\partial\bm{x}_l } \frac{\partial\bm{x}_l}{\partial \bm{x}_{l-1}} =  \left(\nabla_{\bm{x}_l} \mathcal{L} \right)^\top \bm{W}_l, \label{eq:input-gradient}
\end{align}
where $\bm{W}_l$ can be constructed by decompression.
It is worth noting that our \name is compatible with activation checkpointing~\citep{chen2016training} by recomputing the activations, opening more opportunity for memory saving. 

\begin{figure}[t]
\usetikzlibrary{shapes.geometric, arrows.meta, fit, backgrounds, positioning}
\newlength{\ulen}
\setlength{\ulen}{0.613em}

\tikzset{
    permlargeblock/.style = {rectangle, draw, fill=BrickRed!20, 
                    text width=5\ulen, text centered, 
                    minimum height=1.5em, rounded corners=0.2em},
    permsmallblock/.style = {rectangle, draw, fill=BrickRed!20, 
                         text width=3\ulen, text centered, 
                         minimum height=1.5em, rounded corners=0.2em},
    templargeblock/.style = {rectangle, draw, fill=RoyalBlue!20, 
                    text width=5\ulen, text centered, 
                    minimum height=1.5em, rounded corners=0.2em},
    tempsmallblock/.style = {rectangle, draw, fill=RoyalBlue!20, 
                    text width=3\ulen, text centered, 
                    minimum height=1.5em, rounded corners=0.2em},
    permcompblock/.style = {rectangle, draw, fill=BrickRed!20, 
                    text centered, font=\fontsize{5}{5}\selectfont, rounded corners=0.1em, inner xsep=1.5pt,inner ysep=1.5pt},
    permlegend/.style = {rectangle, draw, fill=BrickRed!20, 
                    minimum width=\ulen, minimum height=\ulen, rounded corners=0.01em},
    templegend/.style = {rectangle, draw, fill=RoyalBlue!20, 
                    minimum width=\ulen, minimum height=\ulen, rounded corners=0.01em},
}
    \centering
    \begin{subfigure}[b]{0.22\textwidth}
        \centering
        \begin{tikzpicture}[font=\fontsize{5.5}{5.5}\selectfont, show background rectangle]
            \node [permsmallblock] (input) {Input};
            \node [permlargeblock, above=2\ulen of input] (weight) {Weight};
            \node [permsmallblock, above=2\ulen of weight] (output) {Output};
            \node [tempsmallblock, right=3\ulen of output] (gradout) {Gradient output};
            \node [tempsmallblock, right=3\ulen of input] (gradin) {Gradient input};
            \node [permlargeblock, right=\ulen of weight] (gradweight) {Gradient weight};
        
            \draw [-stealth] (input) -- (weight);
            \draw [-stealth] (weight) -- (output);
            \draw [-stealth] (gradweight) -- (gradin);
            \draw [-stealth] (gradout) -- (gradweight);
            \draw [-stealth] (input) -- (gradweight);
        
            \draw [-stealth] (output.north) -- ++(0, 1.5em);
            \draw [stealth-] (gradout.north) -- ++(0, 1.5em);
            \draw [stealth-] (input.south) -- ++(0, -1.5em);
            \draw [-stealth] (gradin.south) -- ++(0, -1.5em);
        \end{tikzpicture}
        \caption{Vanilla}
        \label{fig:autograd:vanilla}
    \end{subfigure}
    \begin{subfigure}[b]{0.22\textwidth}
        \centering
        \begin{tikzpicture}[font=\fontsize{5.5}{5.5}\selectfont, show background rectangle]
            \node [tempsmallblock] (input) {Input};
            \node [permlargeblock, above=2\ulen of input] (weight) {Weight};
            \node [tempsmallblock, above=2\ulen of weight] (output) {Output};
            \node [tempsmallblock, right=3\ulen of output] (gradout) {Gradient output};
            \node [tempsmallblock, right=3\ulen of input] (gradin) {Gradient input};
            \node [permlargeblock, right=\ulen of weight] (gradweight) {Gradient weight};
        
            \draw [-stealth] (input) -- (weight);
            \draw [-stealth] (weight) -- (output);
            \draw [-stealth] (gradweight) -- (gradin);
            \draw [-stealth] (gradout) -- (gradweight);
            \draw [-stealth] (input) -- (gradweight);
        
            \draw [-stealth] (output.north) -- ++(0, 1.5em);
            \draw [stealth-] (gradout.north) -- ++(0, 1.5em);
            \draw [stealth-] (input.south) -- ++(0, -1.5em);
            \draw [-stealth] (gradin.south) -- ++(0, -1.5em);
        \end{tikzpicture}
        \caption{AC}
        \label{fig:autograd:ac}
    \end{subfigure}
    \begin{subfigure}[b]{0.22\textwidth}
        \centering
        \begin{tikzpicture}[font=\fontsize{5.5}{5.5}\selectfont, show background rectangle]
            \node [tempsmallblock] (input) {Input};
            \node [permlargeblock, above=2\ulen of input] (weight) {Weight};
            \node [tempsmallblock, above=2\ulen of weight] (output) {Output};
            \node [tempsmallblock, right=3\ulen of output] (gradout) {Gradient output};
            \node [tempsmallblock, right=3\ulen of input] (gradin) {Gradient input};
            \node [templargeblock, right=\ulen of weight] (gradweight) {Gradient weight};
        
            \draw [-stealth] (input) -- (weight);
            \draw [-stealth] (weight) -- (output);
            \draw [-stealth] (gradweight) -- (gradin);
            \draw [-stealth] (gradout) -- (gradweight);
            \draw [-stealth] (input) -- (gradweight);
        
            \draw [-stealth] (output.north) -- ++(0, 1.5em);
            \draw [stealth-] (gradout.north) -- ++(0, 1.5em);
            \draw [stealth-] (input.south) -- ++(0, -1.5em);
            \draw [-stealth] (gradin.south) -- ++(0, -1.5em);
        \end{tikzpicture}
        \caption{AC+LOMO}
        \label{fig:autograd:ac-lomo}
    \end{subfigure}
    \begin{subfigure}[b]{0.322\textwidth}
        \centering
        \begin{tikzpicture}[font=\fontsize{5.5}{5.5}\selectfont, show background rectangle]
            \node [tempsmallblock] (input) {Input};
            \node [templargeblock, above=2\ulen of input] (weight) {Weight};
            \node [tempsmallblock, above=2\ulen of weight] (output) {Output};
            \node [tempsmallblock, right=3\ulen of output] (gradout) {Gradient output};
            \node [tempsmallblock, right=3\ulen of input] (gradin) {Gradient input};
            \node [templargeblock, right=\ulen of weight] (gradweight) {Gradient weight};
            \node [permcompblock, left=1.5\ulen of weight] (comp) {Compressed};
        
            \draw [-stealth] (input) -- (weight);
            \draw [-stealth] (weight) -- (output);
            \draw [-stealth] (gradweight) -- (gradin);
            \draw [-stealth] (gradout) -- (gradweight);
            \draw [-stealth] (input) -- (gradweight);
        
            \draw [-stealth] (output.north) -- ++(0, 1.5em);
            \draw [stealth-] (gradout.north) -- ++(0, 1.5em);
            \draw [stealth-] (input.south) -- ++(0, -1.5em);
            \draw [-stealth] (gradin.south) -- ++(0, -1.5em);
            \draw [-stealth] (comp) -- (weight);

            \node [templegend, below right=5.75\ulen and 0.1\ulen of comp.west] (tl) {};
            \node [right=-1pt of tl, text centered] {Layer-wise};
            \node [permlegend, below right=7\ulen and 0.1\ulen of comp.west] (pl) {};
            \node [right=-1pt of pl, text centered] {Global};
        \end{tikzpicture}
        \caption{\name}
        \label{fig:autograd:neuzip}
    \end{subfigure}
    \caption{Reverse-mode automatic differentiation (e.g., back-propagation) with different memory-saving techniques for a linear layer. Blocks colored blue are loaded in memory temporarily for the calculation of this layer, whereas the blocks colored red are always in memory throughout training.}
    \label{fig:autograd}
\end{figure}
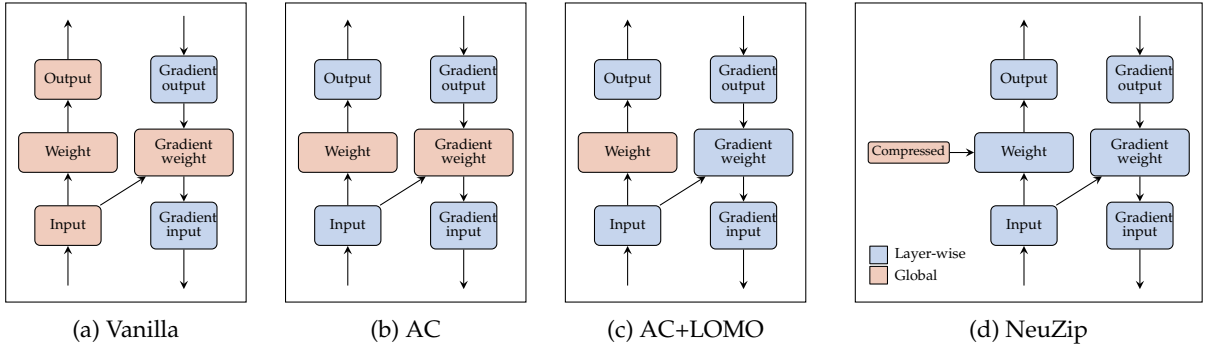

For weight updates, we decompress the matrix into the original floating-point format and compress the updated matrix again. This procedure is done in a layer-by-layer fashion, similar to LOMO~\citep{lv2023full}. The overall training procedure is described in Appendix~\ref{sec:algo}.

\paragraph{Compression algorithm.} In our implementation, we choose to use the asymmetric numeral systems (ANS)~\citep{duda2013asymmetric} as our backbone compression algorithm because it can be easily parallelized and achieves a high throughput with parallel execution, making it an ideal candidate on deep learning accelerators like GPUs. Specifically, ANS encodes a sequence of symbols by treating them as base-$n$ numbers. However, unlike the common numerical system that uses a uniform base for each digit, ANS treats every single digit with a different base $\lceil 1/\hat{p}_i \rceil$, where $\hat{\bm{p}}$ is the frequency of symbols. As a result, it achieves a near-optimal compression rate by suing around $1/\hat{p}_i$ bits for the $i^\mathrm{th}$ symbol.

\subsection{Lossy \name: Additionally Truncating Mantissa for Inference}\label{sec:approach:lossy}

In the algorithm above, we show that the training of neural networks can be completely unaffected by lossless compression. On the other hand, inference is known to be less sensitive to precision loss compared with training~\citep{dettmers2022gptint,dettmers2023case}. This enables further memory reduction of \name by reducing the precision.
In our study, we conduct a pilot experiment that perturbs each weight with a noise proportional to the weight magnitude. We observe that with a small noise ratio there is little or no effect on the overall performance (Appendix~\ref{sec:tolerance}). Motivated by this, we propose a variant of \name that compresses mantissa in a lossy way during inference.

In its core, we simply round and truncate the mantissa to fewer bits. Specifically, we assume the original floating-point number $f$ has an exponent $e$ and mantissa $m$. After rounding, the mantissa is denoted by $\hat m$ and the resulting floating-point number is denoted by $\hat f$.

The rounding introduces an error expressed as: \begin{align}
    | f - \hat{f} | = \left| 2^{e-127} \cdot \frac{m}{2^7}  -  2^{e-127} \cdot \frac{\hat{m}}{2^7}  \right| = 2^{e-134} \cdot |m - \hat{m}|
\end{align}
where $e-127$ interprets the exponent bits $e$ as an integer, which can be either positive, negative, or 0. In the fraction $m/2^7$, $m$ is the significand (an unsigned integer) and $7$ is the precision.
It is straightforward to see that the relative error is given by\begin{align}
    \frac{| f - \hat{f}|}{|f|} = \frac{2^{e-134} \cdot|m - \hat{m}|}{ 2^{e-134} \cdot |m| } = \frac{|m-\hat{m}|}{m}.
\end{align}
Suppose our rounding keeps $k$ most significant bits in $\hat{m}$, the earliest point where $\hat{m}$ could differ from the original number $m$ is at the $(k+1)$th bit. This means that the maximum possible relative change introduced by this rounding is $1/2^k$.
Given that the mantissa bits are highly uniform as shown in Figure~\ref{fig:component-entropies}, such a practice resembles the weight perturbation based on relative magnitudes, justifying the rounding trick applied to mantissas.

In our implementation, we store the sign and mantissa bits together as a signed integer to minimize the requests of memory write. Further, given that modern architectures are mostly byte (8-bit) addressable,
we pack multiple such signed integers into a single byte for memory efficiency. To align with an 8-bit byte, we let the precision after rounding to be $\{0, 1, 3\}$, ensuring that all the bits in a byte are utilized efficiently. We illustrate the process in Figure~\ref{fig:lossy-storage}.

Lastly, we enable a block-wise normalization technique~\citep{dettmers2023qlora}, where a block is a chunk of weights that are stored contiguously in memory. Such block-wise normalization makes sure that the weight with the largest magnitude in a block will always be normalized to 1, invariant to mantissa rounding and truncation. The normalization coefficient---which handles mantissa while ignoring the exponent---is stored with 8 bits, and is used for de-normalization during the decompression of the weight. This strategy is based on the observation that larger weights play a more important role in neural networks~\citep{han2015learning}. 

\section{Experiments}\label{sec:exp}

We empirically verify the effectiveness of \name across different model architectures and datasets. Given the success of large language models, we mainly consider Transformer-based models for our experiments. We choose two designs of Transformer, decoder-only and encoder--decoder models, to show the generality of our method. All experiments are conducted on RTX A6000 GPUs where the uncompressed data type is BFloat16.

\subsection{Lossless \name for Pre-Training}\label{sec:exp:pre-train}

\paragraph{Settings.}  We choose decoder-only models to evaluate our method on the pre-training task. We select 3 models with different sizes to study the scaling effect, including GPT-Neo 2.7B~\citep{gptneo2021black}, Llama-3 8B~\citep{dubey2024llama}, and LLama-2 13B~\citep{touvron2023llama}. For fair comparison, all competing methods are initialized with the same random weights. 

For the task, we consider language modeling, which requires the model to predict the next token given the context. We use the Wikitext-2 dataset~\citep{merity2016pointer}, where each data sample is a fixed-length sequence from an article on Wikipedia. We set the length to 1024 following the common practice~\citep{radford2019language}.

For each experiment, we report the loss (negative log-likelihood) on unseen samples. To study memory saving, we report the peak memory usage for each run during the training process. The numbers are shown in gibibyte (GiB, $1024^3$ bytes). We also report the speed by the number of iterations per second to demonstrate the time-efficiency of each method.

We apply the vanilla SGD update to all runs for efficiency. The activation checkpointing technique~\citep{chen2016training} is enabled by default. It is worth noting that pre-training these large models to the optimal performance is extremely expensive~\citep{rajbhandari2019zero}. Given that our \name training method is lossless, we only train the models for 1 epoch to showcase its effectiveness. We use the same hyper-parameters for all runs.

\paragraph{Results.}
We present the results in Table~\ref{tab:pre-training}. We first test the vanilla training method, where only the activation checkpointing is applied (shown in Figure~\ref{fig:autograd:ac}). As shown, the vanilla training requires the highest amount of memory because it stores the uncompressed weights and gradients for all layers.
 
We also test the LOMO technique~\citep{lv2023full}, which promptly updates the weights in a layer-by-layer fashion (shown in Figure~\ref{fig:autograd:ac-lomo}). This allows LOMO to reuse a buffer to store the gradients for each layer. As a result, LOMO approximately reduces the peak memory usage by the size of a model. 
 
Finally, we apply our \name on top of LOMO (shown in Figure~\ref{fig:autograd:neuzip}). For all models, \name additionally reduces more than 20\% percentage of memory compared with LOMO, accounting for a total memory reduction of more than 50\%. Notably, \name reduces the peak memory of training a Llama-2 13B model to less than 20GB, enabling training a 13B model on consumer-grade GPUs without any precision loss.

\begin{table}[t]
  \caption{Pre-training decoder-only models on the language modeling task. The loss numbers are calculated on the validation set with the cross-entropy loss. Memory is reported in GiB ($1024^3$ B). Speed represents the number of iterations per second. The \textbf{bold} numbers represent the top results.}
  \label{tab:pre-training}
  \centering
  \begin{tabular}{lccccccccc}
    \toprule
    & \multicolumn{3}{c}{GPT-Neo-XL 2.7B } & \multicolumn{3}{c}{Llama-3 8B}    &    \multicolumn{3}{c}{LLama-2 13B}        \\
    \cmidrule(r){2-4} \cmidrule(r){5-7} \cmidrule(r){8-10}
    Name     &  Loss  & Mem & Speed & Loss & Mem & Speed & Loss  & Mem & Speed \\
    \midrule
    Vanilla &  \textbf{8.81} & 11.22 & \textbf{0.96} & \textbf{8.61} & 30.97 & 0.77 & - & OOM & - \\
    LOMO & \textbf{8.81} & 6.97 & 0.94 & \textbf{8.61} & 19.47 & \textbf{0.78}  & \textbf{9.10} & 26.26 & \textbf{0.49} \\
    +\name Lossless & \textbf{8.81} & \textbf{5.54} & 0.70  & \textbf{8.61}& \textbf{15.25} & 0.45  & \textbf{9.10} & 18.58 & 0.28\\
    \bottomrule
  \end{tabular}
\end{table}

\begin{table}[t]
  \caption{Fine-tuning encoder--decoder models on the SQL generation task. The BLEU scores are calculated with SacreBLEU. Memory is reported in GiB ($1024^3$ B). Speed represents the number of iterations per second. The \textbf{bold} numbers represent the top results.}
  \label{tab:fine-tuning}
  \centering
  \begin{tabular}{lccccccccc}
    \toprule
    & \multicolumn{3}{c}{T5 1B} & \multicolumn{3}{c}{T5 3B}    &    \multicolumn{3}{c}{T5 11B}        \\
    \cmidrule(r){2-4} \cmidrule(r){5-7} \cmidrule(r){8-10}
    Name     &  BLEU  & Mem & Speed & BLEU & Mem & Speed & BLEU  & Mem & Speed \\
    \midrule
    Vanilla & \textbf{79.9} & 3.82 & \textbf{3.69} & \textbf{85.1} & 11.32 & 2.43 & - & OOM & - \\                
    LOMO & \textbf{79.9} & 2.75 & 3.68 & \textbf{85.1} & 7.07 & \textbf{2.47} & \textbf{82.3} & 25.95 & \textbf{0.69} \\
    + \name Lossless & \textbf{79.9} & \textbf{2.39} & 2.02 & \textbf{85.1} & \textbf{5.21} & 1.33 & \textbf{82.3} & \textbf{20.68} & 0.46\\
    \midrule
    QLoRA INT8 & 70.4 & 5.84 & 1.11 & 72.1 & 11.54 & 1.12 & 63.5 & 33.36 & 0.37 \\
    QLoRA FP4 & 70.1 &  3.63 & 1.70 & 72.1 & 7.35 & 1.74  & 63.3 & 22.73 & 0.58 \\
    QLoRA FP4$^2$ & 70.6 & 3.61 & 1.63 & 72.0 & 7.27 & 1.61 & 60.6 & 22.38 & 0.57 \\
    QLoRA NF4 & 70.4 &  3.63 & 1.83 & 71.2 & 7.35 & 1.65 & 59.4 & 22.73 & 0.57 \\
    QLoRA NF4$^2$ & 70.5 & 3.61 & 1.64 & 71.2 & 7.07 & 1.57 & 57.9 & 22.38 & 0.57 \\
    \bottomrule
  \end{tabular}
\end{table}

\begin{table}[t]
\caption{Evaluating lossy \name on different models and tasks. `PPL'' represents the perplexity values.  Memory is reported in GiB. Speed represents the number of iterations per second. The \textbf{bold} numbers represent the top results, whereas the \underline{underlined} numbers are the second-best ones.}\label{tab:inf}
\centering

\begin{subtable}[t]{\textwidth}
  \subcaption{Evaluating decoder-only models on the language modeling task. Here, the perplexities are adjusted to word level to compare across different tokenizations. }
  \label{tab:inf:dec}
  \centering
  \resizebox{\linewidth}{!}{
  \begin{tabular}{lccccccccc}
    \toprule
     & \multicolumn{3}{c}{Llama-3 8B} & \multicolumn{3}{c}{Llama-2 13B} & \multicolumn{3}{c}{Yi-1.5 34B}         \\
    \cmidrule(r){2-4} \cmidrule(r){5-7} \cmidrule(r){8-10}
    Name     &  PPL  & Mem & Speed & PPL & Mem & Speed & PPL & Mem & Speed \\
    \midrule
    Vanilla & \textbf{9.89} & 15.08 & \textbf{5.07} & \textbf{10.87} & 24.36 & \textbf{3.59} & - & OOM & - \\
    \midrule
    Quant INT8 & 10.07 & 8.63 & \underline{3.54} & 10.97 & 12.74 & \underline{2.27} & 10.87 & 33.41 & 1.13 \\
    Quant FP4 & 11.51 & 5.77 & 3.45 & 11.38 & 7.37 & 1.87 & 11.57 & 19.54 & \textbf{1.75} \\
    Quant NF4 & 10.75 & 5.77 & 3.38 & 11.15 & 7.37 & 1.83 & 11.06 & 19.54 & \underline{1.67} \\
    Quant FP4$^2$ & 11.50 & \underline{5.44} & 3.41 & 11.38 & \underline{6.87} & 1.86 & 11.57 & \underline{18.11} & 1.61 \\
    Quant NF4$^2$ & 10.75 & \underline{5.44} & 3.34 & 11.15 & \underline{6.87} & 1.81 & 11.06 & \underline{18.11} & 1.54\\
    \midrule
    \name 0-bit & 13.64 & \textbf{5.24} & 3.44 & 12.46 & \textbf{6.30} & 1.87 & 12.06 & \textbf{16.20} & 0.94 \\
    \name 1-bit & 10.77 & 6.05 & 3.38 & 11.17 & 7.77 & 1.86 & 11.04 & 20.14 & 0.93 \\
    \name 3-bit & \underline{9.93} & 7.70 & 3.38 & \underline{10.90} & 10.73 & 1.84 & 10.76 & 27.92 & 0.93 \\
    \name 7-bit (lossless) & \textbf{9.89} & 10.95 & 3.39 & \textbf{10.87} & 16.66 & 1.84 & 10.72 & 43.40 & 0.94\\
    \bottomrule
  \end{tabular}
  }
\end{subtable}

\vspace{.5cm}
\begin{subtable}[t]{\textwidth}
  \subcaption{Evaluating encoder--decoder models on the language modeling task. Since all models use the same tokenizer, we reported perplexities at the token level for simplicity.}
  \label{tab:inf:enc-dec}
  \centering
  \resizebox{\linewidth}{!}{
  \begin{tabular}{lccccccccc}
    \toprule
    &\multicolumn{3}{c}{T5 1B} & \multicolumn{3}{c}{T5 3B}  & \multicolumn{3}{c}{T5 11B}  \\
    \cmidrule(r){2-4} \cmidrule(r){5-7} \cmidrule(r){8-10}
    Name     &  PPL  & Mem & Speed & PPL & Mem & Speed & PPL & Mem & Speed \\
    \midrule
    Vanilla & \textbf{2.614} & 1.37 & \textbf{23.73} & \textbf{2.571} &  5.31 & \textbf{19.86} & \textbf{2.568} & 21.06 & \textbf{6.20} \\
    \midrule
    Quant INT8 & 2.615 & 1.28 & 4.24 & \underline{2.573} & 4.94 & 4.28 & \underline{2.569} & 19.59 & 2.58 \\
    Quant NF4 & 2.632 & 1.08 & 11.64 & 2.588 & 4.12 & 11.82 & 2.579 & 16.28 & 4.48 \\
    Quant FP4 & 2.646 & 1.08 & 11.92 & 2.594 & 4.12 & \underline{11.99} & 2.585 & 16.28 & \underline{4.59} \\
    Quant FP4$^2$ & 2.646 & 1.05 & 10.39 & 2.594 & 4.03 & 9.72 & 2.585 & 15.93 & 4.52 \\
    Quant NF4$^2$ & 2.632 & 1.05 & 10.39 & 2.587 & 4.03 & 9.96 & 2.579 & 15.93 & 4.39 \\
    \midrule
    \name 0-bit & 2.731 & \textbf{0.40} & 11.82 & 2.668 & \textbf{1.41} & 8.70 & 2.651 & \textbf{5.35} & 3.24 \\
    \name 1-bit & 2.641 & \underline{0.48} & 11.68 & 2.591 & \underline{1.78} & 8.61 & 2.581 & \underline{6.65} & 3.21 \\
    \name 3-bit & \textbf{2.614} & 0.66 & \underline{11.99} & 2.574 &2.42 & 8.60 & \underline{2.569} & 9.27 & 3.19 \\
    \name 7-bit (lossless) & \textbf{2.614} & 0.99 & 11.55 & \textbf{2.571} & 3.73 & 8.77 & \textbf{2.568} & 14.46 & 3.23 \\
    \bottomrule
  \end{tabular}
  }
\end{subtable}
\end{table}

\subsection{Lossless \name for Fine-Tuning}\label{sec:exp:fine-tune}

\paragraph{Settings.} A benefit of using lossless compression comes from retaining the pre-trained weight without any information loss. We conduct a fine-tuning experiment with encoder--decoder models to test the performance of our \name on broader architectures. In particular, we choose
three T5 models: T5 1B, T5 3B, and T5 11B~\citep{raffel2020exploring}, where the pre-trained parameters are used for initialization.

The T5 models are pre-trained on the C4 dataset~\citep{lin2020exploring}, which is filtered to contain natural language only. To avoid data leaks from pre-training, we choose a non-natural language generation dataset for fine-tuning. Specifically, we use a public SQL generation dataset~\citep{zhong2017seq2sql,yu2018spider} as the test bed. For each sample, the model is required to generate the SQL command from a human question. For example, the question could be ``\verb|CREATE TABLE head (age INTEGER)|. How many heads of the departments are older than 56 ?''. The model is expected to generate ``\verb|SELECT COUNT(*) FROM head WHERE age > 56|''. We feed the question and response into the encoder and decoder, respectively. The objective is to minimize the cross-entropy loss on the response.

Similar to the pre-training experiments, we also sweep the learning rate from $10^{-3}$ to $3 \times 10^{-1}$ for each run. After fine-tuning, we generate with the model on the validation set with greedy decoding. The generated SQL commands are then compared with the ground truths by SacreBLEU~\citep{post2018call}, a metric that evaluates the similarity between corpora based on precision scores.

\paragraph{Results.}
The results are reported in Table~\ref{tab:fine-tuning}.  All baselines in the pre-training experiment (i.e., the vanilla training, LOMO, and \name) are included in this table. Similar to the results in Section~\ref{sec:exp:pre-train}, they achieve the same BLEU scores for each model. Specifically, our \name is able to train a 11B model within 24GB.

For fine-tuning, it is possible to apply other memory-efficient training techniques. For example, QLoRA~\citep{dettmers2023qlora} compresses the pre-trained model by using low-precision data types and train the LoRA modules only~\citep{hu2022lora}. In our comparison experiment, we choose the widely used quantization data types for QLoRA, including INT8~\citep{dettmers2022gptint}, FP4, and NF4~\citep{dettmers2023qlora}. We apply the LoRA modules~\citep{hu2022lora} on all linear layers, where every LoRA rank is set to $8$ to control the memory usage.\footnote{It should be noted that the down-projection matrices in each T5 feed-forward network are not quantized for stability, as otherwise the model performance is seriously jeopardized. See \url{https://github.com/huggingface/transformers/issues/20287} for more details.} As shown in the second half of Table~\ref{tab:fine-tuning}, all quantization methods underperform \name in terms of both generation quality and memory usage. In terms of time efficiency, some quantization methods are slower than others, but in general, they are in the same magnitude as our method. Overall, \name achieves the least memory usage while maintaining the highest performance.
The results strongly suggests the practicality of our \name.

\subsection{Lossy Compression for Inference}\label{sec:exp:inf}
As mentioned in Section~\ref{sec:approach:lossy}, the inference process is less sensitive in precision loss, which provides an opportunity for compressing mantissa in a lossy fashion during inference. We evaluate the performance of our lossy \name in such scenarios.

\paragraph{Settings.} Following the settings in previous sections, we test our approach with both decoder-only and encoder--decoder architectures. For the decoder-only models, we select the LLama-3 8B~\citep{dubey2024llama}, LLama-2 13B~\citep{touvron2023llama}, and Yi-1.5 34B~\citep{ai2024yi}. For the encoder--decoder architecture, we use the T5 1B, 3B, and 11B models as in Section~\ref{sec:exp:fine-tune}.

Since all decoder-only models are trained for language modeling, we evaluate the performance with language modeling tasks. Specifically, we test all methods on the Wikitext-2 validation set~\citep{merity2016pointer} following Section~\ref{sec:exp:pre-train}, where each sequence consists of 1024 tokens.  On the other hand, the encoder--decoder models (T5 series) contain multiple tasks in pre-training. Since they excel at zero-shot translation, we evaluate them on the WMT14 En-De translation task~\citep{wmt14ende}, where each source sentence is prepended with ``translate from English to German:'' based on the pre-training format~\citep{raffel2020exploring}.

Following the standard evaluation pipeline for lossy compression~\citep{frantar2023optq,dettmers2023case}, we evaluate all models with the perplexity metric, which is sensitive to how distorted the compressed model is.

\paragraph{Results.}
The results for decoder-only and encoder--decoder models are shown in Tables~\ref{tab:inf:dec} and \ref{tab:inf:enc-dec}, respectively. 
We see that the vanilla (uncompressed BFloat16) models achieve the best perplexity scores in all experiments at a cost of the excessive memory usage.
For quantization methods, we choose the same INT8~\citep{dettmers2022gptint}, FP4, and NF4~\citep{dettmers2023qlora} data types mentioned in Section~\ref{sec:exp:fine-tune}. In general, quantization methods suffer from notable perplexity degradation. Although the INT8 variant~\citep{dettmers2022gptint} manages to better preserve the perplexity, it uses around 50\% more memory compared with other quantization methods.

For our lossy \name, we set three different levels of precision: 0-bit, 1-bit, and 3-bit mantissa preserved. We choose these values because they are aligned in 8-bit byte arrays (discussed in Section~\ref{sec:approach:lossy}).  All these variants use a block size of 512 for normalization. We additionally include the lossless \name (7-bit mantissa) for a full comparison.
As shown in the table, our lossy \name demonstrates a spectrum of memory saving and performance preservation. The 0-bit \name attains the best memory efficiency in all experiments, whereas the lossless 7-bit \name obtains the best perplexity scores. Notably, the 3-bit \name achieves nearly lossless performance in all experiments while using less than 50\% memory compared with the uncompressed model. The results confirm the effectiveness of our method.

 \begin{figure}[!t]
    \centering
    \includegraphics[width=\textwidth] {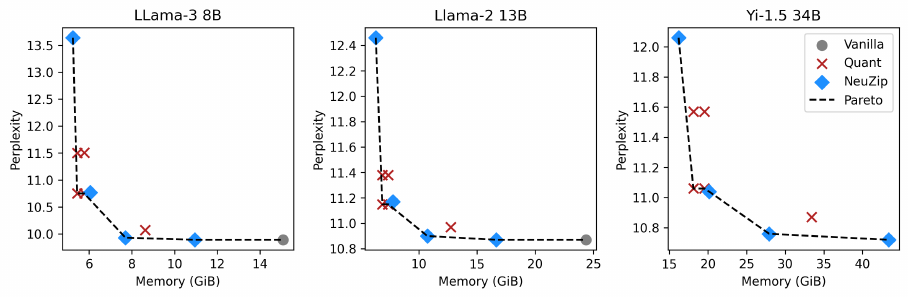}\vspace{-.2cm}
    \caption{The trade-off between memory and performance for different methods.}\vspace{-.2cm}
    \label{fig:pareto}
\end{figure}

\subsection{In-Depth Analyses}\label{sec:exp:analyses}
 \paragraph{The memory--performance trade-off.} In Section~\ref{sec:exp:inf}, we observe that the performance is generally decreased with less memory usage. We analyze this trade-off of our \name as well as quantization methods in Figure~\ref{fig:pareto}. Note that the optimal methods are the ones on the Pareto frontier~\citep{pareto2014manual}, i.e., the more bottom-left, the better. In addition to measuring the perplexity, we also include a preliminary study by evaluating the end-to-end performance on the MMLU dataset~\citep{hendrycks2020measuring} in Appendix~\ref{sec:apx:mmlu}.

As shown, three out of four \name variants are on the Pareto frontier, with the remaining one staying fairly close to the frontier. On the other hand, there is only one quantization technique that lies on the Pareto frontier. This result  demonstrates that our \name generally achieves a better memory--performance trade-off than quantization.

\paragraph{The effect of block size in lossy compression.} As introduced in Section~\ref{sec:approach}, we apply normalization to lossy \name to ensure the weight with the largest absolute value will not be affected by truncation. We show the effect of block size in this experiment with a giant model, Llama-3 70B evaluated on the Wikitext-2 dataset.

\begin{table}[!t]
  \caption{The effect of block size.}
  \label{tab:block-size}
  \centering
  \resizebox{0.95\linewidth}{!}{
  \begin{tabular}{lcccccccccc}
    \toprule
 &\multicolumn{2}{c}{Block 32} &\multicolumn{2}{c}{Block 64} &\multicolumn{2}{c}{Block 128} &\multicolumn{2}{c}{Block 256} &\multicolumn{2}{c}{Block 512} \\
   \cmidrule(r){2-3} \cmidrule(r){4-5} \cmidrule(r){6-7} \cmidrule(r){8-9} \cmidrule(r){10-11}
    Name     &  PPL  & Mem  & PPL & Mem  & PPL & Mem &  PPL & Mem & PPL & Mem  \\
    \name 0-bit & 6.341 & 35.7 & 6.694 & 34.6 & 6.853 & 34.2 & 7.639 & 33.8 & 7.104 & 33.5 \\
    \name 1-bit & - & OOM & 4.611 & 42.7 & 4.662 & 42.2 & 4.640 & 41.8 & 4.649 & 41.4 \\
    \bottomrule
  \end{tabular}
  }
\end{table}

As seen in Table~\ref{tab:block-size}, a smaller block size clearly leads to better performance at the cost of compromising memory efficiency due to the overhead of storing normalization coefficients. Therefore, the block-wise normalization provides a more fine-grained trade-off between memory and performance by varying the block size. 

\paragraph{Throughputs of \name.}  

\begin{figure}[h]
    \centering
    \includegraphics[width=0.8\textwidth]{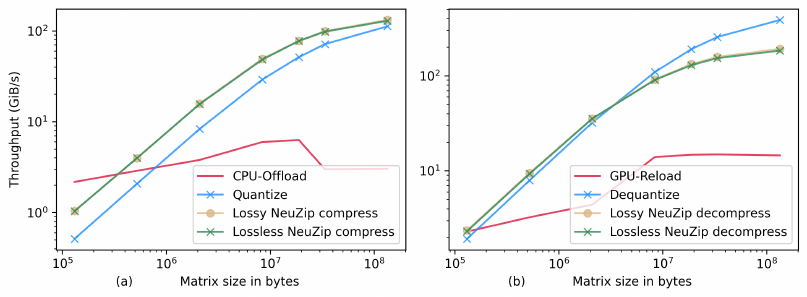}
    \caption{The throughput experiment. (a) Comparison of CPU-offloading, quantization, lossy \name compression, and lossless \name compression. (b) Comparison of GPU-reloading, de-quantization, lossy \name decompression, and lossless \name decompression.}
    \label{fig:throughput}
\end{figure}

In addition to overall time efficiency presented in Tables~\ref{tab:pre-training}--\ref{tab:inf}, we analyze the throughput of matrix compression and decompression with our \name, in comparison with the throughput of matrix quantization and de-quantization based on the NF4 data type~\citep{dettmers2023qlora} using the popular library \texttt{bitsandbytes}.\footnote{Available at \url{https://github.com/bitsandbytes-foundation/bitsandbytes}}
We additionally include the CPU-offloading technique as a baseline, which lowers the GPU memory pressure by transferring data to CPU and reloading them to GPU when needed.
Figure~\ref{fig:throughput} measures the throughput of matrix processing in GiB/s when we vary the matrix size from $10^5$ to $10^8$ bytes.

We see that CPU-offloading is generally slow across different sizes of matrices. This is due to the bottleneck of CPU--GPU communication through PCIe.
For quantization, the \texttt{bitsandbytes} package has been highly optimized for GPU, and its throughput is one magnitude higher than the CPU-offloading technique when the matrix size is large.
Profoundly, our \name achieves the highest throughput for compression among all methods (Figure 4a),  and a high throughput for decompression similar to de-quantization (Figure 4b). The results suggest that our \name, albeit causing overhead compared with uncompressed vanilla models, is still highly efficient in practice.

\section{Related Work}\label{sec:related-work}

\paragraph{Model compression.} Previous work has explored different techniques to reduce the memory usage of neural networks, including knowledge distillation~\citep{hinton2015distilling} and pruning~\citep{kown2022fast}. Most related to our work is the quantization technique, which represents each parameter with fewer bits; common approaches include $k$-means-based quantization~\citep{han2016deep}, linear quantization~\citep{han2016deep}, and mixed precision quantization~\citep{dettmers2022gptint,dettmers2023qlora}. When training data are available, one may incorporate the  quantization into the training process to improve performance~\citep{xiao2023smoothquant,frantar2023optq}. In this paper, our \name compression is a zero-shot method, and therefore, our experiments consider the widely used zero-shot quantization methods~\citep{dettmers2022gptint,dettmers2023qlora} for fair comparison.  We leave the utilization of additional data of \name to future work.

\paragraph{Memory-efficient optimizers.} The optimizer also occupies a considerable amount of memory during training~\citep{rajbhandari2019zero}. To address this, memory-efficient optimizers~\citep{shazeer2018adafactor,zhao2024galore,hao2024flora} are developed to reduce the memory footprint of training. Our \name is orthogonal to these optimization techniques, as it can be seamlessly combined with any of these methods for further memory saving. In particular, the lossless \name is expected to have exactly the same results with less memory.

\paragraph{Parameter-efficient training.} Another line of research saves memory by training a subset of parameters~\citep{houlsby2019parameterefficient,zaken2022bitfit} so the optimizer only stores information about a small set of trainable parameters. One notable example is the low-rank adaptation (LoRA~\citep{hu2022lora}). However, such a practice restricts the optimization space of parameters, and thus usually leads to significant performance degradation. Moreover, low-rank methods are unsuitable for pre-training.

It is important to mention that memory-efficient optimizers and parameter-efficient training cannot reduce the memory cost during inference. By contrast, our \name is suitable for both training and inference.

\section{Conclusion}\label{sec:conclusion}

\paragraph{Summary.} In this work, we present \name, a novel compression scheme for neural networks that achieves memory-efficient training and inference. By analyzing the floating-point structures, we propose to compress the exponent in a lossless way and to compress the mantissa in a lossy way. The lossless variant of our \name may be applied to both training and inference, while yielding exactly the same result as the uncompressed model. The lossy \name provides additional memory saving for inference, achieving superior memory--performance trade-off. 

\paragraph{Limitations and future work.} Due to the hardware constraint, the largest model that we consider in this paper is 70B. We would like to verify our \name on even larger models like GPT-3~\citep{brown2020language} with more capable hardware. Another limitation of \name is that the throughput is lower than the vanilla model. However, it has a comparable speed with highly optimized quantization methods while achieving significantly better performance. By using \name, we expect to create opportunities for researchers with academic budgets to explore and study large models.

\section*{Acknowledgments}
The research is supported in part by the Natural Sciences and Engineering Research Council of Canada (NSERC), a Mitacs Accelerate project, the Amii Fellow Program, the Canada CIFAR AI Chair Program, an Alberta Innovates Program, and the Digital Research Alliance of Canada (alliancecan.ca).

\bibliography{main}

\vfill
\pagebreak

\appendix

\section{Inspecting the Entropy on More Models}\label{sec:apx:more_ent}

\paragraph{Random initialization.}
When training from scratch, the parameters are randomly initialized. To verify the compressibility in this case, we check the parameter entropy of a randomly initialized model with the same architecture as Lllama-3. The initialization methods follow the standard procedure provided in the Hugging Face library. The results show a similar pattern to what the released Llama model has, suggesting the compressibility with \name occurs even with random weights.

\begin{figure}[h]
    \centering
    \includegraphics[width=1\linewidth]{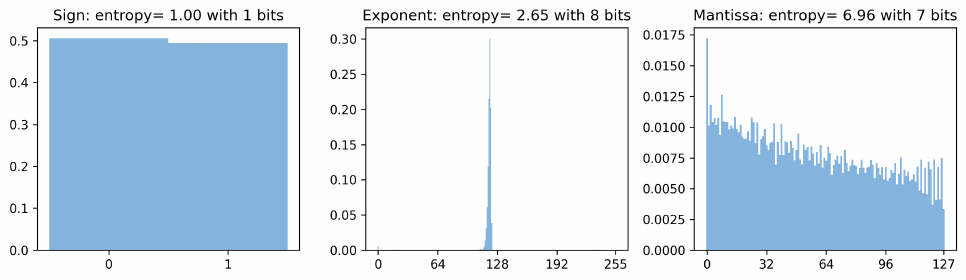}
    \caption{The histograms of different floating-point components of the parameters of a randomly initialized Llama-3 8B model. }
    \label{fig:random}
\end{figure}

\paragraph{Diffusion.}
We also inspect the parameter entropies beyond Transformer models. In Figure~\ref{fig:diffusion}, we check all four models in a diffusion pipeline. We see that the low-entropy exponents not only occur in Transformer models but other architectures like convolution-based VAE and U-Net models.

\begin{figure}[h]
    \centering
    \includegraphics[width=1\linewidth]{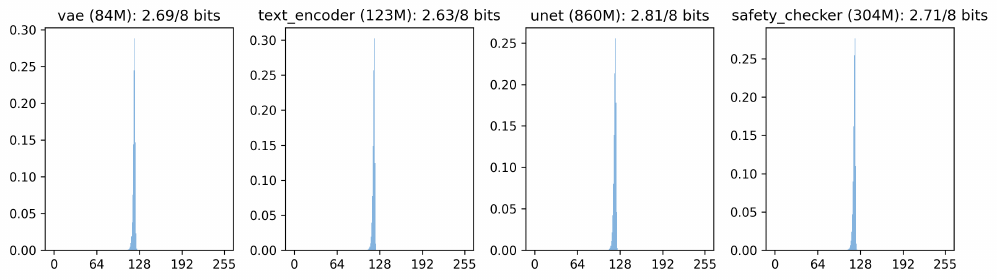}
    \caption{The histograms of the exponent bits in different components of Stable Diffusion 1.5 model. We omit the sign and mantissa bits for simplicity as we do not compress them based on their entropies. }
    \label{fig:diffusion}
\end{figure}

Both experiments show that the occurrence of low-entropy components is a common phenomenon in deep learning.

\section{The Algorithm for Training with Lossless \name}\label{sec:algo}

 In this section, we describe the forward-backward procedure of \name. First, we compress all the linear layers in the original model and store the compressed information on-device. During the training iterations, we decompress the compressed weights in a layer-by-layer manner for the forward pass. For the backward pass, the input is recalculated again following the forward pass like activation checkpointing~\citep{chen2016training}. A linear operation calculates the gradients for both the weight matrix and input. To do so, we need to decompress the weight again, which is used to calculate the gradient of input. After the gradient is calculated, we directly update the weight without storing it similar to LOMO~\citep{lv2023full}.
 
\begin{algorithm}[h]
    \small
    \caption{Memory-efficient training with \name}
    \label{alg:training}
 \begin{algorithmic}[1]
     \REQUIRE number of linear layers $L$, linear layer weights $\{\bm{W}_i\}_{i=1}^L$.
     \REQUIRE data stream $\mathcal D$ that yields training data $\bm{x}$ for each iteration

     \COMMENT{Initialization}
     \FOR{$l \in 1 \dots L$}
        \STATE{$\bm{s}_l, \bm{e}_l, \bm{m}_l \gets \mathrm{split}(\bm{W}_l)$} \hfill \COMMENT{Split each element in the matrix into three components}
        \STATE{$\bm{c}_l \gets \mathrm{compression}(\bm{e}_l)$} \hfill \COMMENT{Compress the exponents losslessly}
        \STATE{$\mathrm{store}(\bm{s}_l, \bm{c}_l, \bm{m}_l)$} \hfill \COMMENT{Store the compressed exponents $\bm{c}_L$ on device}
     \ENDFOR
     
     \COMMENT{Training loop}
    
    \FOR{$\bm{x}$ in $\mathcal D$}
        \STATE{}\COMMENT{Model forward}
        \STATE {$\bm{x}_0 \gets \bm{x}$} 
         \FOR{$l \in 1 \dots L$}
               \STATE{$\hat{\bm{e}} \gets \mathrm{decompression}(\bm{c}_l)$} \hfill \COMMENT{Decompress the exponents using temporary space}
               \STATE{$\hat{\bm{W}} \gets \mathrm{merge}(\bm{s}_l, \hat{\bm{e}}_l, \bm{m}_l)$} \hfill \COMMENT{Concatenate into a floating-point number matrix using temporary space}
               \STATE {$\bm{x}_{l} \gets \hat{\bm{W}}^\top\bm{x}_{l-1} + \bm{b}_l$} \hfill \COMMENT{Linear calculation}
               \STATE{$\mathrm{save\_for\_backward}(\bm{x}_l)$} \hfill \COMMENT{Label the variable required for back-propagation}
         \ENDFOR

         \STATE{}\COMMENT{Model backward and update}
         
         \STATE{$\Delta_{\bm{x}} \gets {\partial \mathcal L / \partial \bm{x}_L}$} \hfill \COMMENT{Calculate the gradient w.r.t. the model output}
         \FOR{$l \in L \dots 1$}
               \STATE{$\hat{\bm{e}} \gets \mathrm{decompression}(\bm{c}_l)$} \hfill \COMMENT{Decompress the exponents using temporary space}
               \STATE{$\hat{\bm{W}} \gets \mathrm{merge}(\bm{s}_l, \hat{\bm{e}}_l, \bm{m}_l)$} \hfill \COMMENT{Concatenate into a floating-point number matrix using temporary space}
              \STATE{$\Delta_{\bm{W}} \gets (\Delta_{\bm{x}}) \bm{x}_{l-1}^\top$} \hfill \COMMENT{Calculate gradient by Equation~\eqref{eq:linear-gradient}}
              \STATE{$\hat{\bm{W}} \gets \hat{\bm{W}} - \mathrm{optimizer}(\Delta_{\bm{W}})$} \hfill \COMMENT{Update the weight on-the-fly}
                  \STATE{$\bm{s}_l, \bm{e}_l, \bm{m}_l \gets \mathrm{split}(\hat{\bm{W}})$} \hfill \COMMENT{Split each element in the matrix into three components again}
              \STATE{$\bm{c}_l \gets \mathrm{compression}(\bm{e}_i)$} \hfill \COMMENT{Compress the exponents losslessly again}
              \STATE{$\mathrm{store}(\bm{s}_l, \bm{c}_l, \bm{m}_l)$} \hfill \COMMENT{Replace the stored components for layer $l$ on device}
              \STATE{$\Delta_{\bm{x}} \gets \hat{\bm{W}} \Delta_{\bm{x}}$} \hfill \COMMENT{Calculate the gradient of input for next layers}
         \ENDFOR
     \ENDFOR
 \end{algorithmic}
 \end{algorithm}

 \section{The Tolerance of Random Perturbation}\label{sec:tolerance}

 In this experiment, we would like to explore the sensitivity of neural network weights to random perturbations. For each parameter, we have two types of magnitudes: absolute and relative magnitudes. The former one represents the actual numerical error, whereas the second one is calculated based on the original value. For example, when the original value is $-1.5$, an absolute magnitude of $0.125$ means the perturbed range is $[-1.5 - 0.125, -1.5 + 0.125]$. On the other hand, a relative magnitude of $0.125$ means the perturbed range is $[-1.5 * (1+0.125), -1.5 * (1-0.125]$.
We conduct such a experiment with the perturbation grid in Figure~\ref{fig:relative-error}. For each cell, we choose the maximum error between relative error and absolute value for perturbing. A random value is  sampled from the perturbed range uniformly as the perturbation. The weight value is then set to the random sample.

 \begin{figure}[h]
    \centering
    \includegraphics[width=0.8\textwidth]{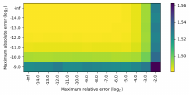}
    \caption{Evaluating the byte-level perplexity with perturbed LLama-3 8B model~\citep{dubey2024llama} on Wikitext-2~\citep{merity2016pointer}. Each parameter is perturbed with controlled noises. Both the $x$- and $y$-axes are log-scale with base 2.}
    \label{fig:relative-error}
\end{figure}

As shown in Figure~\ref{fig:relative-error}, we see a clear pattern that the model tends to tolerate the relative change rather than the absolute change.

\section{The Storage for Lossless and Lossy Compression}\label{sec:storage}
In this section, we describe the underlying storage layout for \name in Figure~\ref{fig:storage}.

\begin{figure}[h]
    \begin{subfigure}[h]{0.47\textwidth}
      \centering
      \includegraphics[width=\textwidth]{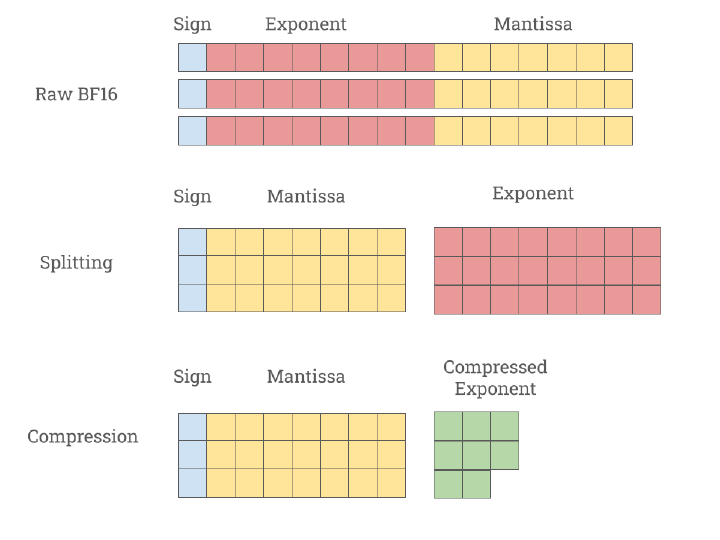}
      \subcaption{Lossless compression scheme.}
      \label{fig:lossless-storage}
  \end{subfigure}
  \begin{subfigure}[h]{0.47\textwidth}
      \centering
      \includegraphics[width=\textwidth]{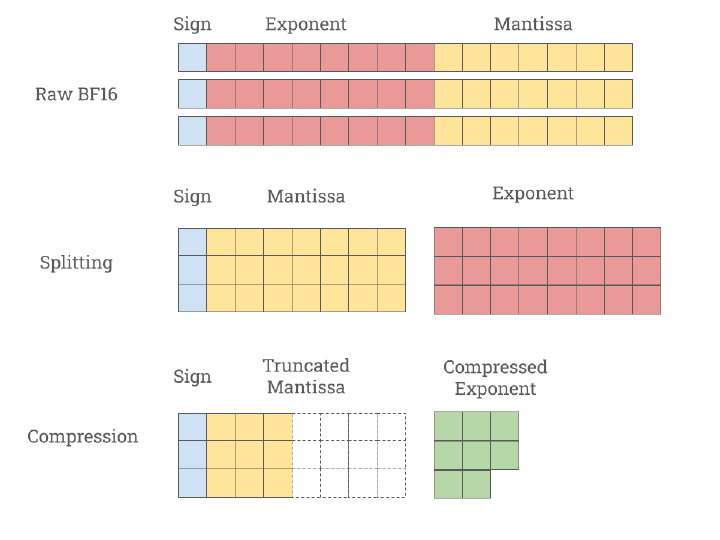}
      \subcaption{Lossy compression scheme (with 3 bits).}
      \label{fig:lossy-storage}
  \end{subfigure}
  \caption{The storage structures for \name.}
  \label{fig:storage}
  \end{figure}

Essentially, each BFloat16 number is first split into an exponent and a signed mantissa. We group all the exponents in the matrix and perform the lossless compression. The signed mantissa is optionally truncated, depending on the required precision. The signed mantissa is then stored separately in memory.

\section{Evaluating on MMLU}\label{sec:apx:mmlu}

We provide the results on MMLU~\citep{hendrycks2020measuring} in Figure~\ref{fig:mmlu}. Here, the theoretical optimal point should be at the top left corner.

\begin{figure}[H]
    \centering
    \includegraphics[width=0.5\linewidth]{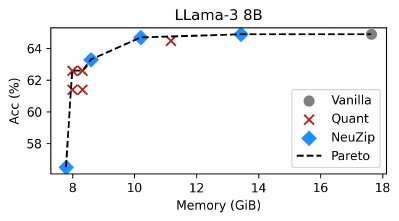}
    \caption{The memory--performance trade-off on the MMLU dataset.}
    \label{fig:mmlu}
\end{figure}

Similar to the results in Section~\ref{sec:exp:analyses}, all of our \name variants are on the Pareto frontier, suggesting the optimal trade-off between memory and performance.

\vfill
\pagebreak

\end{document}